\documentclass{article}
\usepackage{spconf,amsmath,graphicx}
\usepackage{booktabs}
\usepackage{float}
\usepackage[utf8]{inputenc}
\usepackage[T1]{fontenc}
\usepackage{amssymb}
\usepackage{array}
\usepackage{multirow}
\usepackage{lipsum}
\usepackage{graphicx}
\usepackage{placeins}
\usepackage[colorlinks]{hyperref}  %

\newcommand{\etal}{\textit{et al.\hspace{0.5mm}}}

\title{Anomaly Object Segmentation with \\
Vision-Language Models for Steel Scrap Recycling}
\name{Anonymous ICIP 2025 Submission}
\address{Paper ID \#1500}

\name{Daichi Tanaka, Takumi Karasawa, 
Shu Takenouchi, Rei Kawakami}
\address{Institute Science of Tokyo}
\begin{document}
%
\maketitle
\begin{abstract}
Recycling steel scrap can reduce carbon dioxide (CO2) emissions from the steel industry. However, a significant challenge in steel scrap recycling is the inclusion of impurities other than steel. To address this issue, we propose vision-language-model-based anomaly detection where a model is finetuned in a supervised manner, enabling it to handle niche objects effectively. This model enables automated detection of anomalies at a fine-grained level within steel scrap. Specifically, we finetune the image encoder, equipped with multi-scale mechanism and text prompts aligned with both normal and anomaly images. The finetuning process trains these modules using a multiclass classification as the supervision. 
In terms of average precision (AP) at the pixel level, our framework achieved 28.6\%, whereas existing studies, constrained by their frozen image encoder, achieve less than 1.0\%.
\end{abstract}
%
%
\section{Introduction}
\label{sec:intro}
Steel is produced by forming crude steel, which emits a large amount of carbon dioxide (CO2). The steel industry accounts for 7 to 9\% of global CO2 emissions (2020) \cite{worldsteel}. Greenhouse gases, including CO2, significantly contribute to global warming, necessitating emission reductions.
The recycling of steel scrap has gained attention as a solution. Steel scrap is an important material for addressing challenges in the steel industry, such as resource conservation and CO2 emissions reduction, and recycling it can reduce CO2 emissions by approximately 975Mt \cite{worldsteel2024_blog}.

While the importance of recycling steel scrap has increased, many challenges remain in the recycling process. Steel scrap may contain contaminants and other metals. If these impurities are mixed in, it 
leads to a reduction in the strength of the manufactured products and causes further CO2 emissions, making anomaly detection crucial.
In addition, anomaly detection is now 
performed
by workers through visual inspection, leading to variations in accuracy and inefficiencies. Therefore, automatic detection of anomalies at the region level in steel scrap 
is urgently needed.

Image-based anomaly detection 
includes reconstruction-based approaches utilizing autoencoders\,\cite{choi2023subspaceprojectionapproachautoencoderbased}, GANs\,\cite{akcay2018ganomaly, Akcay2019SkipGANomaly}, and diffusion models \cite{hu2024anomalydiffusionfewshotanomalyimage}. These methods identify anomaly 
by analyzing reconstruction failures, but
they face a fundamental challenge: the model, trained extensively on normal data, must fail when encountering anomalies.
Another group of methods relies on pretrained models to detect anomaly regions by identifying features that deviate from those of normal images \cite{10.1007/978-3-030-68799-1_35, Roth_2022_CVPR}. Separately, few-shot approaches, which often utilize vision-language foundation models (VLMs) or general foundation models \cite{cao_segment_2023, chen2023zero, Jeong_2023_CVPR}, incorporate prior knowledge into the detection process by leveraging text descriptions of anomalies.

\begin{figure}[t]
    \centering
    \includegraphics[width=1\linewidth]{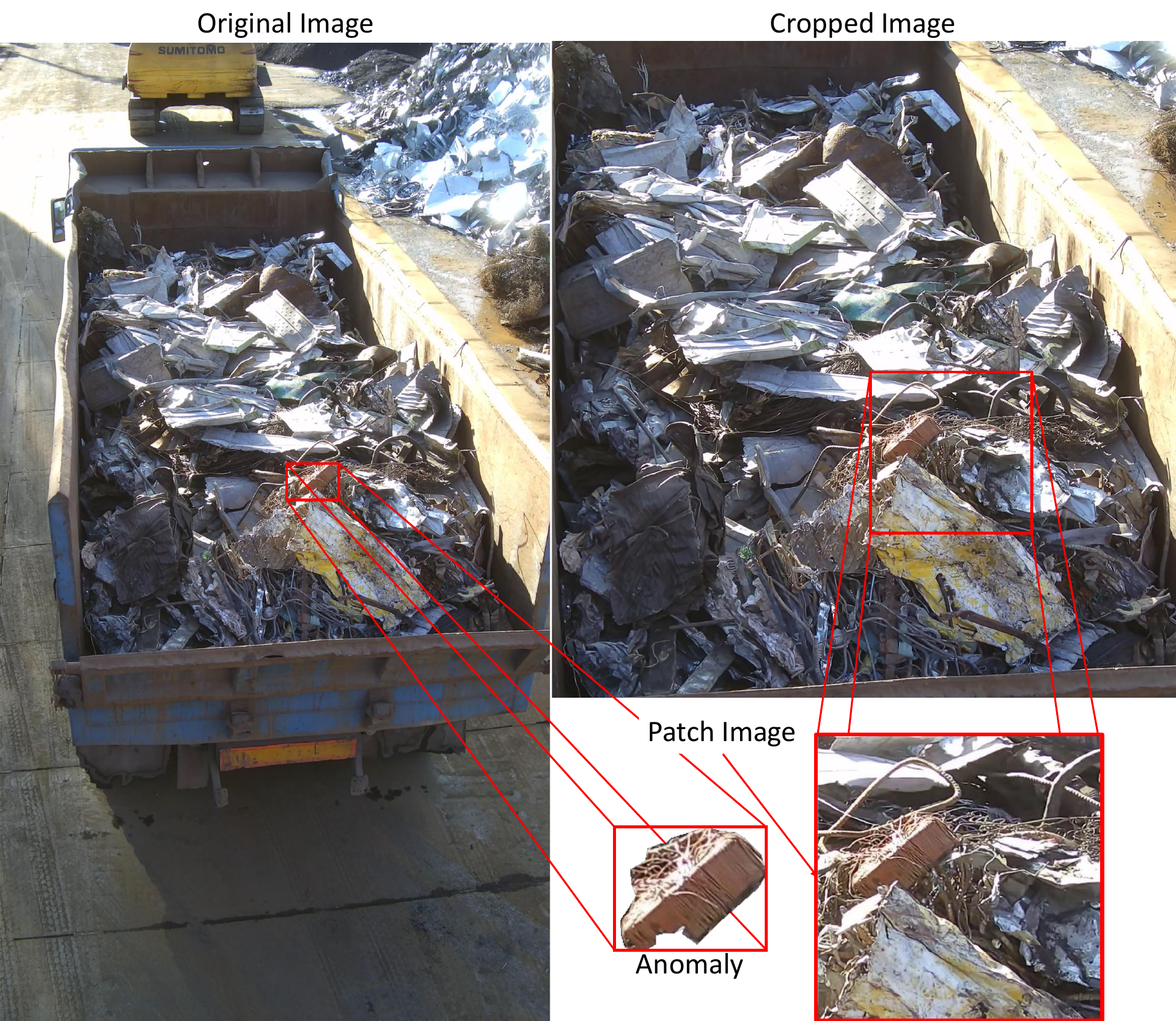}
    \vspace{-5mm}
    \caption{An example of the steel scrap dataset. The original image has a 4K resolution. The cropped image is obtained by extracting the truck bed portion from the original image using Language Segment-Anything \cite{medeiros2024langsegmentanything}. Its size varies depending on the image. The patch image is a fixed-size segment cut from the cropped image, set to 512×512 in the experiment. This anomaly belongs to the \textit{motor} class.}
    \label{fig1:my_label1}
\end{figure}

However, these methods cannot be applied directly to the problem addressed in this study, due to the following three issues. The first challenge is the data specificity. Compared to general anomaly detection datasets such as MVTec \cite{8954181}, VisA \cite{10.1007/978-3-031-20056-4_23} and MVTec LOCO AD \cite{Bergmann2022}, steel scrap images are niche. Furthermore, the anomalies in steel scrap vary widely in appearance and characteristics. Even for humans, it is difficult to visually distinguish between normal and anomaly items without specialized knowledge.
The second is the data scarcity. This is a common challenge in anomaly detection, as anomalies are scarce, and there is an imbalance in the number of labeled data.
The third is the clutter of the images. 
The shapes of the objects, both normal and anomaly, are diverse due to their inherent nature, as well as their tendency to overlap and be partially occluded, as shown in Fig.\,\ref{fig1:my_label1}. Consequently, distinguishing between the foreground (anomaly) and the background (steel scrap) is more challenging than in standard anomaly detection datasets \cite{8954181, 10.1007/978-3-031-20056-4_23, Bergmann2022}.

To address these issues, we propose a CLIP \cite{pmlr-v139-radford21a}-based anomaly detection framework where a vision encoder is finetuned in a supervised manner.
VLMs \cite{pmlr-v139-radford21a, pmlr-v162-li22n, pmlr-v202-li23q} can leverage knowledge in text, and CLIP \cite{pmlr-v139-radford21a} has a text encoder that can extract latent text information that can be a guide to detect anomalies from visual features.
Along with the prompt tuning of text features, we let the image encoder of CLIP \cite{pmlr-v139-radford21a} undergo 
supervised training using hundreds of anomaly images. This training is necessary since the images are domain specific, and the model needs to perform multiclass classification to separate normal and multiple types of anomaly objects. A multi-scale mechanism is also introduced to extract feature representations at different scales.


In the experiment, we tested our framework as well as the selected baselines, WinCLIP \cite{Jeong_2023_CVPR} and AnomalyCLIP \cite{zhou2023anomalyclip}, on our steel scrap dataset. 
An example image and cropped patch is shown in Fig.\,\ref{fig1:my_label1}. In terms of average precision (AP) at the pixel level, our framework achieved 28.6\%, whereas existing methods, constrained by their frozen image encoders, achieved less than 1.0\%. This study not only clarifies the anomaly detection performance for steel scrap recycling with CLIP, but also shows insights in cases where domain-specific knowledge is required for image-based anomaly detection.


\vspace{-1mm}
\section{Related work}
\label{sec:format}
\vspace{-1mm}
\noindent \textbf{Anomaly detection with foundation models} \hspace{1mm}
Since the advent of vision foundation models \cite{pmlr-v139-radford21a, pmlr-v162-li22n, pmlr-v202-li23q, Kirillov_2023_ICCV}, 
these models have been explored for anomaly detection. 
In particular, CLIP \cite{pmlr-v139-radford21a} is often utilized due to its aligned feature space for text and images. 
For instance, Jeong \etal
\cite{Jeong_2023_CVPR} and Chen \etal \cite{chen2023zero} utilize the similarity between text features and multi-scale image features based on CLIP for anomaly detection.
These methods fix both the image and text encoders; thus, they struggle with niche objects, such as those in our dataset, which lie beyond the scope of the pretrained knowledge.

To overcome these limitations, 
recent studies 
refine text prompts and incorporate patch-level feature extraction, such as the approach proposed by Zhou \etal \cite{zhou2023anomalyclip}. These methods refine the text features through prompt tuning so that the features become more anomaly-specific. However, 
the image features from a frozen encoder
are still limited when dealing with a diverse range of anomaly patterns or when both normal and anomaly objects exhibit a wide variety of characteristics, as is the case with steel scrap in our dataset.

Our framework not only finetunes text prompts but also trains the CLIP \cite{pmlr-v139-radford21a} image encoder via supervised learning, because specialized knowledge has to be incorporated to handle niche objects effectively.

\vspace{1mm}
\noindent \textbf{Image recognition in steel scrap classification}\hspace{1mm}
For completeness, we review image recognition techniques introduced in steel scrap recycling. Most studies focus on steel grade classification, as it is just as important as anomaly detection.

Daigo \etal \cite{IchiroDaigo2023ISIJINT-2022-331}
enhances the accuracy of semantic segmentation for steel scrap with complex shapes using multi-scale feature extraction. 
Zhong \etal \cite{math12091370} also utilizes multi-scale feature extraction and employs wavelet-based down-sampling and up-sampling to better preserve features.
Tu \etal \cite{9893213} proposes a comprehensive framework where the truck bed is first segmented, then parts of steel scrap are detected, and the grade level is estimated based on the detection results.   
All these methods utilize a number of real-world steel scrap images with labels for intensive supervised model training.

Inspired by these studies,
we also employ multi-scale feature extraction and train our model on steel scrap images with labels. However, our focus is on anomaly detection, where annotated data is often insufficient for training.
%
We therefore leverage the
vision-language foundation models \cite{pmlr-v139-radford21a} (VLMs), which can mitigate data scarcity problem through extensive pretraining.

\begin{figure*}[tb]
    \centering
    \includegraphics[width=1\linewidth]{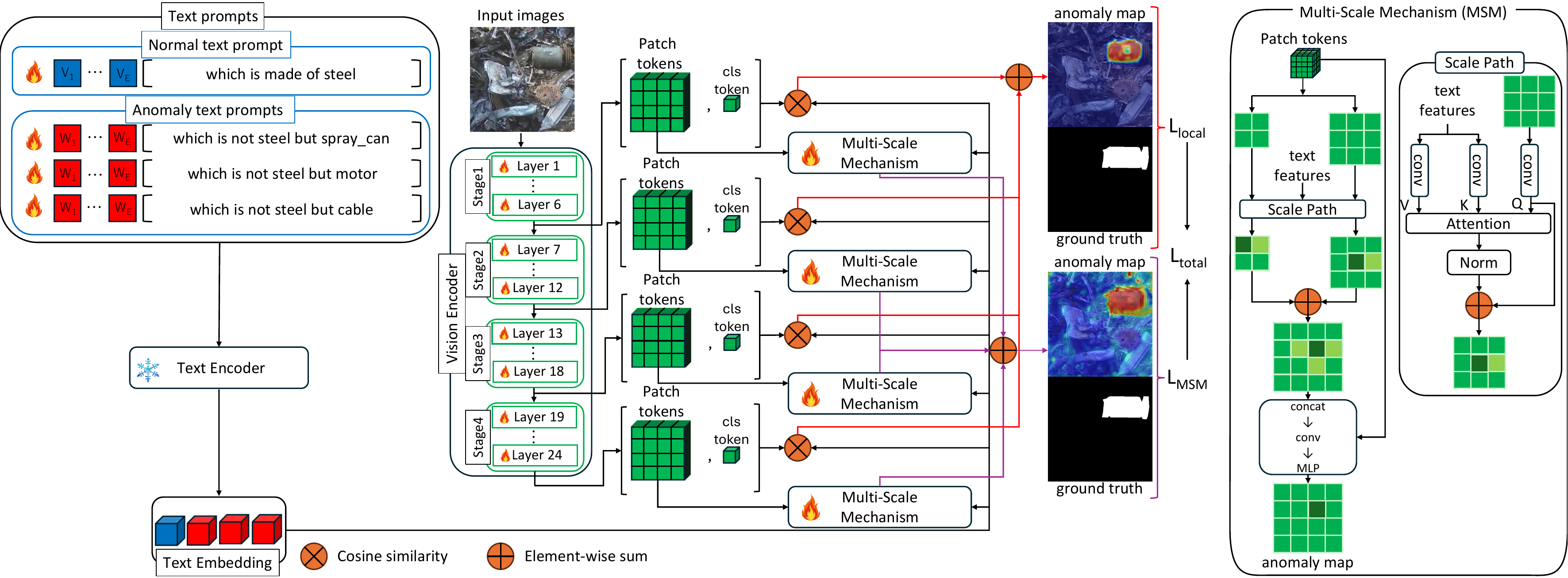}
    \vspace{-4mm}
    \caption{The overall process of the anomaly detection based on a vision-language model \cite{pmlr-v139-radford21a}. In our approach, text prompts, the vision encoder and the multi-scale mechanism (MSM) are trained using supervised learning. To acquire knowledge of niche objects, we employ a multiclass classification as a loss function. The image on the right illustrates the structure of MSM. MSM extracts patch features at multiple scales, enabling it to capture the contours of overlapping or occluded objects, as well as the features of objects of varying sizes.}
    \label{fig2:my_label2}
\end{figure*}

\vspace{-1mm}
\section{Method}
\label{sec:pagestyle}
\vspace{-1mm}
We have three main challenges: the uniqueness of the data, data scarcity, and the cluttered nature of the images.
To address these challenges, we propose an anomaly detection model based on a  VLM. 
VLMs \cite{pmlr-v139-radford21a,pmlr-v162-li22n, pmlr-v202-li23q}, pretrained on hundreds of millions of images, offer the ability to address data scarcity with their vast knowledge base. 
Larger models are better few-shot learners; thus, finetuning of the visual encoder of them requires a small amount of images to be able to differentiate anomalies in these niche objects.
VLMs also demonstrate generalization capabilities for unseen objects through the integration of image and text information; thus, 
VLMs 
enables accurate identification of anomalies
with the aid of provided text information.

Among various VLMs \cite{pmlr-v139-radford21a, pmlr-v162-li22n, pmlr-v202-li23q}, we choose CLIP \cite{pmlr-v139-radford21a} for two reasons. The first is because of its aligned latent space between text and image features.
CLIP \cite{pmlr-v139-radford21a}, pretrained on image-text pairs, can detect parts of images with features similar to those of given text descriptions. 
The second is its simple architecture. It has independent image and text encoders, which simplify the anomaly detection model. 

In our framework, the CLIP's image encoder and text prompts are finetuned in a supervised manner using both normal and anomaly images. This finetuning allows CLIP to acquire specialized knowledge about niche objects. 
Adjusting text prompts enhances utilization of text space for our anomalies.
In this supervision, a multi-class classification is employed rather than 
the binary classification of anomaly and normal objects.
By distinguishing features of different anomaly types, the model avoids confusion.
To account for the variability in object sizes within our dataset, we introduce a multi-scale mechanism (MSM) for extracting and integrating multi-scale features,
inspired by the success of previous work \cite{pspnet,LI2025129122}, particularly ClipSAM\cite{LI2025129122}. 
MSM captures features of varying sizes while preserving spatial information. 

The anomaly detection model 
is illustrated in the left side of Fig.\,\ref{fig2:my_label2}. The utilization of text information to find anomalous regions is inspired by the architecture of AnomalyCLIP \cite{zhou2023anomalyclip}. The text prompts, encoded by the frozen text encoder, are used to compute the similarity between image and those texts.
Here, text prompts are prepared for each type of anomaly for multiclass learning. The initial prompts are as follows: the normal prompt is ``object which is made of steel'' and the anomaly prompts are ``object which is not steel but \{anomaly object's name\}'' as depicted in Fig.\,\ref{fig2:my_label2}.

The image features are extracted from a vision transformer \cite{dosovitskiy2021an}, which has four stages. Features are obtained at each stage as the image passes through them.
During this process, tokens of each patch and the CLS token, a token that represents the entire image, can be obtained. For each patch token, we compute the similarity with text prompts, and the similarities from each stage are summed to create a single anomaly map for each anomaly object.

The patch features also go through the MSM. 
This process results in another anomaly map that includes multi-scale spatial information. The MSM is shown on the right in Fig.\,\ref{fig2:my_label2}. It applies average pooling with different strides to create tokens that contain information at coarser scales. Those tokens go through a scale path that attends to anomalous regions identified by the text information at these scales. The scale path produces psuedo anomaly maps with multiple scales and they are  up-sampled to be converted back to the original resolution of the patch tokens. Psuedo anomaly maps at multiple scales are summed and encoded to produce a final anomaly map. MSM is introduced in each stage of the vision encoder.

The resulting anomaly maps are compared with ground truth. For each anomaly map, losses are calculated using two loss functions, 
multiclass classification with data imbalance and the segmentation loss. 
The specific losses we used are provided in the experimental section. 

As a pre-process, we use Language Segment-Anything \cite{medeiros2024langsegmentanything} to create cropped images of the truck bed from the original anomaly images. Then, they are further divided  into patch images. An example of a cropped image is shown in Fig.\,\ref{fig1:my_label1}.
\section{Experiment}
\label{sec:typestyle}
\subsection{Setup}
\begin{figure}[tb]
    \centering
    \includegraphics[width=.9\linewidth]{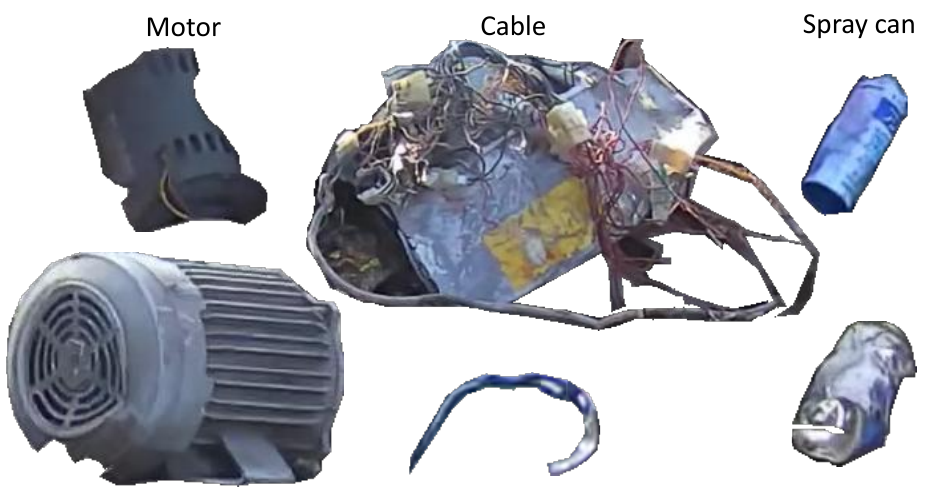}
    \vspace{-1mm}
    \caption{An example of anomaly objects cropped from the anomaly images used in our experiment.}
    \label{fig3:my_label3}
\end{figure}

\noindent \textbf{Dataset and Metrics.} \hspace{1mm}
The images used in this study were collected from a steel scrap recycling site and provided by 
\textit{anonymized}.
The images were captured using a 4K camera mounted on a crane that transports the steel scrap. Only frames in which the entire truck was correctly captured were manually selected. A total of 6,754 images were collected, encompassing various types of anomaly images. We constructed a dataset focusing on three categories with a large amount of data provided: motor, cable, and spray can.
Examples of anomaly objects are presented in Fig.\,\ref{fig3:my_label3}.
The original anomaly images, one of those examples is shown in Fig.\,\ref{fig1:my_label1}, and the text prompt \textit{the scrap} were input into Language Segment-Anything \cite{medeiros2024langsegmentanything}. We split the cropped images as shown in Fig.\,\ref{fig1:my_label1}, and they were further divided into 512$\times$512 patches. The ground truth for the anomaly images was generated in a similar manner. The breakdown of the dataset is presented in Table.\,\ref{tab1:dataset_breakdown1}.

For evaluation, we employed four metrics—Area Under the Receiver Operating Characteristic Curve (AUROC), Average Precision (AP), AUPRO, and F1-Max—to assess the performance of local anomaly detection at the pixel level. These metrics are widely used in previous studies, including WinCLIP \cite{Jeong_2023_CVPR} and AnomalyCLIP \cite{zhou2023anomalyclip}.

\begin{table}[t]
\centering
\caption{Breakdown of the dataset used for training and testing in our experiment. Each number represents the number of images. During testing, a binary classification approach is used for evaluation, where motor, cable, spray can are classified as anomalies.}
\vspace{2mm}
\begin{tabular}{l|c|ccc}
    \toprule
    \multirow{2}{*}{\textbf{}} & \multirow{2}{*}{\textbf{Normal}} & \multicolumn{3}{c}{\textbf{Anomaly}} \\ 
                                       &                                 & \textbf{Motor} & \textbf{Cable} & \textbf{Spray Can} \\ 
    \midrule
    \textbf{Train}                    & 7838                            & 467            & 172            & 97 \\ 
    \midrule
    \textbf{Test}                     & 2065                            & \multicolumn{3}{c}{353} \\ 
    \bottomrule
\end{tabular}
\label{tab1:dataset_breakdown1}
\end{table}

\vspace{1mm}
\noindent \textbf{Anomaly detection model.} \hspace{1mm}
CLIP \cite{pmlr-v139-radford21a} image encoder employed in this study was based on the vision transformer \cite{dosovitskiy2021an} (ViT)-L/14@336px architecture. The text prompt, image encoder, and multi-scale mechanism were all trained, with a learning rate of 1.0×$10^{-6}$ using the schedule-free \cite{defazio2024road} variants of AdamW \cite{Loshchilov2017DecoupledWD}.
Other parameters included 15 epochs, an input image size of 518, and feature extraction from layers 6, 12, 18, and 24 for calculating similarity with text prompts. The batch size was 4. In the multi-scale mechanism (MSM), we used 3$\times$3 and 9$\times$9 scales. During testing, binary classification was performed.
The weights for the three anomaly prompts were averaged.

For text prompt finetuning, the parameters were set as follows: the learnable text prompt had a length of 16, a depth of 9, and an embedding length of 4.
Furthermore, the initial prompts for multi-class classification were described in Sec. 3. For binary classification, the normal prompt remained the same as in multiclass classification, and the anomaly prompt was set as ``object which is not steel.''

The loss functions were the class-balanced focal loss \cite{Cui_2019_CVPR} and dice loss \cite{li-etal-2020-dice} for classification and segmentation, both adapted for multiclass version. 
Using these two loss functions, we defined \(L_{\text{local}} \) as the loss between the anomaly map, obtained from the similarity between patch features and text prompts, and the ground truth.  \( L_{\text{MSM}} \) represents the loss between the anomaly map obtained from the MSM module
and the ground truth.
\( L_{\text{local}} \) was computed as the sum of the losses across all stages. In contrast, \( L_{\text{MSM}} \) was calculated by weighting the losses from stages 1 to 3 by 0.1 and the loss from stage 4 by 0.7, before summing them.

\begin{figure*}[tb]
    \centering
    \includegraphics[width=1.0\linewidth]{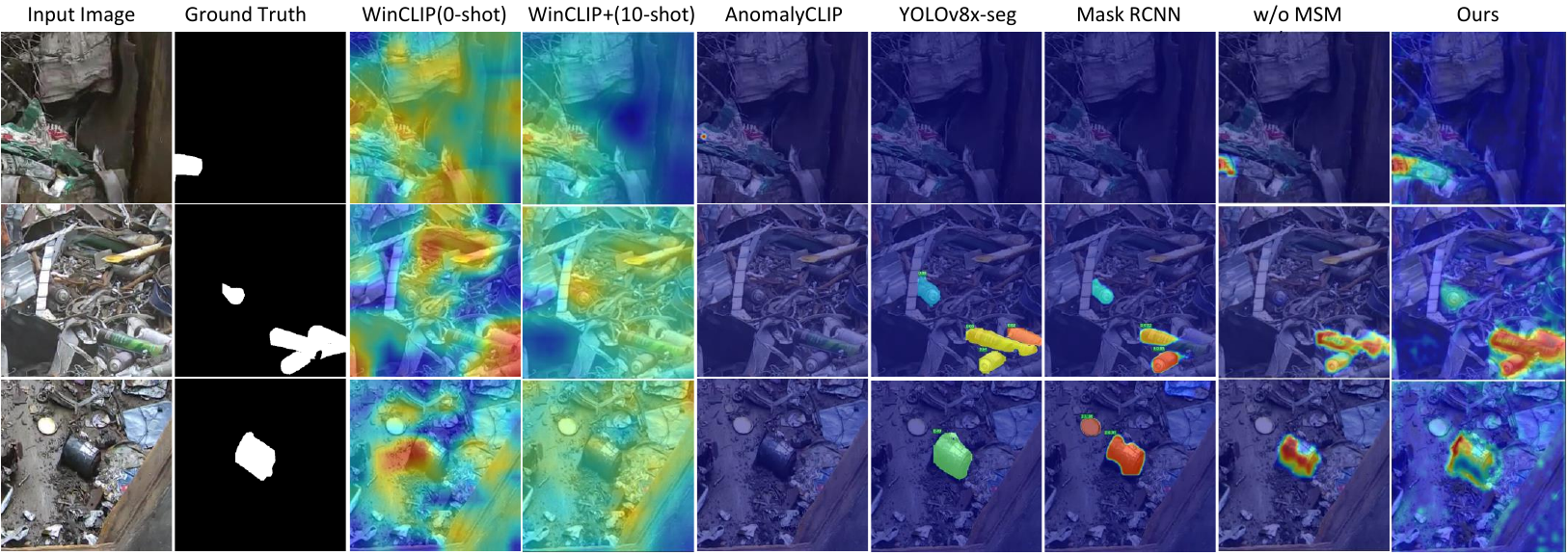}
    \vspace{-3mm}
    \caption{Qualitative comparison between existing methods, and ours with and without MSM. Ours detects anomaly surrounded by the steel scrap more accurately.\\[3pt]}
    \label{fig4:my_label4}
\end{figure*}

\subsection{Results}
\noindent \textbf{Comparison with baseline methods.} \hspace{1mm}
We compared ours against YOLOv8 \cite{yolov8_ultralytics2023} combined with SAM (Segment Anything Model) \cite{Kirillov_2023_ICCV}, YOLOv11 \cite{yolo11_ultralytics} combined with SAM, and Mask R-CNN \cite{he2017mask}.
For YOLOv8, and YOLOv11, we used the size of extra large. For Mask R-CNN, we used ResNet50.  We finetuned these models with the data we used for training, which includes the class of anomalies and the annotation mask. For YOLO models, we combined the output with SAM to obtain the segmentation mask.

We also compared ours with WinCLIP \cite{Jeong_2023_CVPR}, WinCLIP+ \cite{Jeong_2023_CVPR}, and AnomalyCLIP \cite{zhou2023anomalyclip}.
This comparision is to show the limitation of the generalizability of zero/few-shot anomaly detection models on our dataset; thus, the training data were limited for these models. For WinCLIP \cite{Jeong_2023_CVPR}, a zero-shot model, only the prompts were provided to the model to fit the target domain of this study. For WinCLIP+ \cite{Jeong_2023_CVPR}, not only the prompts but also the ten normal images from the training dataset were provided to the model as reference images.
For AnomalyCLIP \cite{zhou2023anomalyclip}, which learns prompts during training, the initial prompts were set to ``object which is made of steel'' for normal objects and ``object which is not steel'' for anomaly objects. AnomalyCLIP as well as our model were trained on our dataset, while AnomalyCLIP performed prompt-tuning of anomalies and normals (binary), while our model performed multi-class prompt-tuning and learning of the vision encoder as in Fig.\ref{fig2:my_label2}. 
Our results were the averages of three experiments.

The results are shown in Table.\,\ref{tab2:test_results2}.
Ours consistently outperformed all existing approaches across all evaluation metrics. Notably, the average precision, which is robust to data imbalance, exhibited a dramatic improvement—from 10.29\% by the Mask R-CNN to 28.6\% in our approach. This demonstrates 
the power of learning a vision encoder and a text prompt
to be able to detect anomaly objects in steel scrap recycling, which are very domain-specific.

Fig.\,\ref{fig4:my_label4} presents qualitative results. WinCLIP+ \cite{Jeong_2023_CVPR} and AnomalyCLIP \cite{zhou2023anomalyclip} cannot detect the locations of anomalies as those models do not learn visual cues. 
Finetuned models such as YOLO and Mask R-CNN can detect anomalies; however, they struggle with challenging cases—such as the one shown in the top row—where the anomalies closely resemble normal objects.
In contrast, our method accurately detects anomalies at more precise locations, which is visually evident. These results demonstrate that our method is well-suited for anomaly detection in steel scrap.

\begin{table}[tb]
\centering
\vspace{-2mm}
\caption{Comparison with baseline methods. Ours results were the averages of three experiments.\\[3pt]}
\vspace{-4mm}
\begin{tabular}{l|>{\centering\arraybackslash}p{8mm}>{\centering\arraybackslash}p{8mm}>{\centering\arraybackslash}p{8mm}>{\centering\arraybackslash}p{8mm}}
    \toprule
    \textbf{Method} & \textbf{auroc (\%)} & \textbf{aupro (\%)} & \textbf{ap (\%)} & \textbf{f1max (\%)} \\
    \midrule
    \multirow{1}{*}{WinCLIP \cite{Jeong_2023_CVPR}} & 50.54 & 6.44 & 0.06 & 0.14 \\ 
    \multirow{1}{*}{WinCLIP+ \cite{Jeong_2023_CVPR}(10shots)} & 52.58 & 7.18 & 0.07 & 0.21 \\ 
    \multirow{1}{*}{AnomalyCLIP \cite{zhou2023anomalyclip}} & 50.4 & 0.6 & 0.4 & 1.2 \\ 
    \midrule
     \multirow{1}{*}{YOLOv8-seg} & 63.4 & 10.73 & 6.93 & 17.18 \\
     \multirow{1}{*}{YOLOv11-seg} & 60.3 & 8.08 & 4.36 & 15.07 \\     
     \multirow{1}{*}{Mask R-CNN} & 66.3 & \textbf{20.46} & 10.29 & 22.64 \\          
    \multirow{1}{*}{Ours ($\frac{\Sigma x}{3}$)} & 
    \textbf{85.4} & 19.7 & \textbf{28.6} & \textbf{35.9} \\
    \bottomrule
\end{tabular}
\label{tab2:test_results2}
\end{table}

\vspace{1mm}
\noindent \textbf{Multi-scale mechanism ablation} \hspace{1mm}
The impact of scale size in the multi-scale mechanism (MSM) and the presence or absence of multi-scale on performance was evaluated using combinations of (3$\times$3 \& 9$\times$9), (5$\times$5 \& 11$\times$11), and (7$\times$7 \& 13$\times$13). Three experiments were conducted for each metric, and average values were recorded.

Table.\,\ref{tab3:test_results3} summarizes the effect of MSM. The last row presents the results without MSM.
When MSM is introduced, the performance generally improves. The combination of (3$\times$3 \& 9$\times$9) shows the highest performance across all metrics. This may due to the size distribution of anomaly objects. 
The size of anomalies has a long-tail distribution, and
the sizes between 3$\times$3 and 9$\times$9 were particularly common. 
The combination of (5$\times$5 \& 11$\times$11) did not perform as good as other combinations, which we consider it may due to the stochasticity of the learning. Qualitative results in Fig.\,\ref{fig4:my_label4} shows that anomalies were detected with MSM, with the spray can in the center of the image being detected only with MSM. This indicates that MSM enhances anomaly detection by extracting features from diverse object sizes.


\vspace{-1mm}
\begin{table}[t]
\centering
\vspace{-2mm}
\caption{Evaluation of the effect of scale size and the presence of the multi-scale mechanism on anomaly detection performance. The results were the averages of three experiments.}
\begin{tabular}{l|>{\centering\arraybackslash}p{8mm}>{\centering\arraybackslash}p{8mm}>{\centering\arraybackslash}p{8mm}>{\centering\arraybackslash}p{8mm}}
    \toprule
        \textbf{Scale sizes} & \textbf{auroc (\%)} & \textbf{aupro (\%)} & \textbf{ap (\%)} & \textbf{f1max (\%)} \\
    \midrule
    \multirow{1}{*}{3×3 \& 9×9 ($\frac{\Sigma x}{3}$)} & 85.4 & 19.7 & 28.6 & 35.9 \\ 
    \midrule
    \multirow{1}{*}{5×5 \& 11×11 ($\frac{\Sigma x}{3}$)} & 83.8 & 18.5 & 25.9 & 33.6 \\ 
    \midrule
    \multirow{1}{*}{7×7 \& 13×13 ($\frac{\Sigma x}{3}$)} & 85.7 & 19.8 & 27.1 & 34.1 \\ 
    \midrule
    \multirow{1}{*}{w/o MSM ($\frac{\Sigma x}{3}$)} & 83.5 & 16.5 & 25.3 & 33.3 \\ 
    \bottomrule
\end{tabular}
\label{tab3:test_results3}
\end{table}

\section{Conclusion}
\label{sec:majhead}
We have presented a framework combining an anomaly detection model based on CLIP \cite{pmlr-v139-radford21a}, one of the vision-language foundation models \cite{pmlr-v162-li22n, pmlr-v202-li23q} (VLMs) for application in steel scrap recycling.
We can detect and localize the anomaly objects in steel scrap 
more effectively than both finetuned object detection models with segmentation capabilities and zero-shot VLM-based models.
We demonstrated the high adaptability of VLMs for anomaly detection in scrap metal recycling.


\begin{small}
\bibliographystyle{IEEEbib}
\bibliography{refs}
\end{small}

\end{document}